# Effects of Dataset properties on the training of Generative Adversarial Networks

Ilya Kamenshchikov, ikamenshchikov@gmail.com,
Matthias Krauledat, matthias.krauledat@hochschule-rhein-waal.de
Hochschule Rhein-Waal

*Abstract* – Generative Adversarial Networks are a new family of generative models, frequently used for generating photorealistic images. The theory promises for the GAN to eventually reach an equilibrium where generator produces pictures indistinguishable for the training set. In practice, however, a range of problems frequently prevents the system from reaching this equilibrium, with training not progressing ahead due to instabilities or mode collapse. This paper describes a series of experiments trying to identify patterns in regard to the effect of the training set on the dynamics and eventual outcome of the training.

1. INTRODUCTION

## 1.1. Generating Images

Generating images is a task with many applications. As images are a compact and convenient format for communicating for humans, it is desirable for a computer to be able to generate such, as this would enable users to understand a wide range of messages and information faster and with ease. While there exist multiple software tools for generating images, for example photoshop, they are merely a way for a human to translate their idea into an image and take significant amount of effort and experience. There also exist abundant amount of 3D engines, many of them written for computer games. 3D engines can render objects as seen from any angle and distance. Although some families of objects can have their 3d models generated programmatically, these are usually non-complex objects or parts of the scene background. More complex objects, such as characters, monsters, weapons, etc., are drawn by human professionals for use in those 3D engines.

While some graphics are based around 3d models, eventually they all are displayed as 2D graphics on the computer screen. For this reason, as well as higher complexity in 3D, we will concentrate on the task of generating 2D images in this work.

The fundamental difficulty with generating 2D images comes from the fact that in the entire space of images, ones that would be described as noise take up great part of the space. Indeed, randomly sampling the images per pixel has virtually no chance of resulting in an image of a real object. Therefore, the model generating images must possess the principles according to which a "real" image can be generated. Such rules come in many cases from the nature of 3d space and objects in it. This involves many complex rules, such as perspective, light reflection, and many others. One approach would be to try to program these rules into the software. This is basically what most of 3D engines do – they simulate the rules of the real world around us in hope to create a realistic rendering in this way. While efficient in creating stunning visual effects, this does not lead to the software understanding the second part of creating new images – their content and logic.

Machine Learning offers a tool for a model to extract the rules of the 2D image space from the data, which nowadays is becoming ever more abundant. When having a model learn to create pictures, we care for two parts – the image must be realistic according to the rules of the domain and look visually attractive; and the content of the image must make sense. Some failed attempts show that it is indeed not rare to have good visual quality while having the content that makes little sense. Such images look disturbing to a human, e.g. 3 eyed dogs.

## 1.2. Metrics for comparing a generative model

A successful generative model should ideally possess all of the following qualities:

1. Creating images of high visual quality. This can be measured subjectively by a human trying to answer a question "could this be a real picture". There also exist some metrics that can be used by computer and resulting in a number, as will be discussed in methods section. [1]

2. Diversity – we would like a model to be able to produce a wide variety of images. Having a model that produces only one image all the time is not worth much, even if the image is of great visual quality.

3. Control over the content and / or style of the generated images – we would like the model to produce specific images on demand. This is also called having latent space of images

– for example for photos of a person, one might wish to be able to ask for specific hair color, eye color, male or female picture, etc.

**1.3. Machine Learning models for generating images**

There are not too many families of successful ML models capable of generating images. The two most known and widely recognized are Autoencoders and Generative Adversarial Networks (GAN).

**Autoencoders** take the approach of trying to restore an image after compressing it to its latent space. The model must compress an image to a small vector of real numbers (say, 100 numbers). Then it must restore the image only from this information and achieve something which is as close as possible to the original.

The difficulty with Autoencoders come from the fact that during a normal training machine learning model will see batches of very different images. In trying to find an approach that makes an improvement on the whole batch or dataset, the model inevitably converges to the mode where blurry images are created, although they are generally correct reconstructions of the original. This way model indeed minimizes the usual training objective, as typical loss function is frequently formulated as per-pixel distance from the original image.

**Generative Adversarial Networks (GAN)** use completely different loss function, which overcomes the problem of blurry images trying to match entire batches at once. In this approach, two networks are trained. One, called Generator, is tasked with producing images. The second one, Discriminator, is tasked with telling which images are real and which are "fake" (generated). The discriminator in this way must acquire the knowledge of the structure of real images. The trick of this approach is that Generator can optimize towards fooling the discriminator – its task is to produce such images that Discriminator will with highest probability judge the image to be coming from the real dataset.

Both Generator and Discriminator are usually constructed as Convolutional Neural Networks, which are widely recognized as the most efficient models for work with 2D data.

Next, we discuss the formulation of loss functions for both discriminator and generator in the Generative Adversarial Network.

**Discriminator**

Loss formulation for the discriminator is based on the classification task it is designed to fulfill. The coming images will be coming from two distributions - real images $R$ and fake images $F$ (those created by the generator network).. We train the discriminator to classify them into two classes by outputting the probability that image $x$ comes from the real images $R$:

$D(x) = p(x \text{ comes from } D)$

The truth about image x is encoded in the label $y$, with $y = 1$ for images from $R$ and $y = 0$ for images coming from $F$. The loss is binary cross entropy, also known as logloss:

$L(D(x) | x, y) = -y * log(D(x)) - (1-y) * log(1 - D(x))$ (1)

Observing that $y$ takes only discrete values 1 and 0, we can verify the semantics of this loss function.

For real images $x$ from $R$ we have $y = 1$:

$y = 1 \rightarrow 1 - y = 0 \rightarrow L(D(x) | x, y=1) = - log(D(x))$.

This function approaches zero as the outputted probability of the image belonging to $R$ approaches 1.

For fake images $x$ from $F$ we have $y = 0$:

$y = 0 \rightarrow 1-y = 1 \rightarrow L(D(x) | x, y= 0) = - log (1 - D(x))$

Observing that 1 - D(x) is the probability of the image being fake, we can see that this formulation works for both cases. In real training we will have multiple images in the same batch summing up their losses towards a common average, but for each single one of them only one part of the equation will be non-zero. This loss function is minimized by discriminator outputting numbers close to 1 for real data and 0 for generated images. It is also a smooth function on the interval (0,1) allowing for the well-conditioned gradients.

**Generator**

The generator works in that it receives a multidimensional noise vector $z$ (which can be roughly understood as an encoding of the future content of the image), and produces an image $x$:

$G(z) = x$

It does so by applying multiple non-linear transformations to the initial vector, which usually increase its dimensionality. In the case of convolutional neural network, these transformations are parameterized by the weights and biases of the layers of the neural network. To have it train to full the discriminator, we just take the same loss function, but with negative sign. At generator training, naturally only generator weights can be changed, which in turn make x to be such an image for which the discriminator is fooled.

$L(G(z) | D, z) = - L (D (x | x = G(z)), y = 0)$

since again y = 0 removes one of the loss terms, it becomes simply:

$L(G(z) | D, z) = log (1 - D(G(z)) )$ (2)

The generator loss is minimized by producing images for which discriminator outputs probabilities of the image being real close to 1.

Training regime: As the loss functions of Generator and Discriminator are contradicting each other (one grows as the other one decreases), the most straightforward approach to training the entire GAN is to let Discriminator and Generator take learning steps in turns (one after another, through the entire training).

As the task of creating images appears in practice to be more difficult that to determine which image is generated, most problems for this family of models arise from discriminator getting too strong, and therefore leaving generator no significant way to improve. In this case, the visual quality of created pictures will be low, as well as having the content in images that is not consistent.

The second possible problem may arise from generator creating only one kind of picture all the time, or a little number of different images. This problem is called mode collapse.

If measures are taken to overcome the above-mentioned difficulties, or the dataset is not particularly complex, the equilibrium of such models is reached in that generator is producing images that are indistinguishable from the ones coming from the dataset, and discriminator is reduced to producing 0.5 for any image, as it can not tell the difference anymore.

Such equilibrium means that the model produces images that are of high visual quality, and diverse. One requirement that is so far left behind is having control over what images are being produced by the model.

**ACGAN**

To gain control over what image will be produced, one may feed additional information to the generator, for example a label for desired image. A model that functions in this manner was published in [2] – Auxiliary Condition GAN (ACGAN).

In this model, the discriminator has not only to tell fake and real pictures apart but be also able to classify the real images according to some classes. The loss part relevant to the classification is the same to standard supervised learning.

The discriminator network will now output two values:

scalar $y$ - probability of the data coming from $R$, vector $c$ - probability of the example $x$ belonging to each of classes with length equal to the amount of classes $n\_classes$ in the real dataset.

The discriminator loss becomes a sum of the binary cross entropy loss for the label y determining if the image is real or fake, and multiclass cross entropy H for classification of real images. For fake images, we use the same loss as in original formulation (eq1). We denote the true label of the example $x$ as $c'$, being a one-hot vector (having one on the position of the correct class and zeros elsewhere). The two outputs of the discriminator are denoted $D_{adv}$ for the probability of images being fake (forming basis for the adversarial loss) and $D_c$ for the vector of probabilities for each of the class labels. $H$ is then the new auxiliary loss factor used to enable generation of specific classes.

$$H(c \mid x, c') = - \sum_{i=0}^{n} (c'_i * \log(c_i) + (1 - c'_i) * \log(1 - c_i)) \quad (3)$$

H is added both to the adversarial discriminator loss and generator loss, making total loss for discriminator be the sum of loss in (1) and auxiliary loss $H$:

$$L_{ACGAN}(D(x) \mid y, c) = L(D_{adv}(x) \mid y) + y * H(D_c(x) \mid c) \quad (4)$$

Generator will be given a label to create each image, in the same format as the discriminator during its training. The task of the generator will be not only to have its image classified as a real one, but also to be given the same label as the one passed during image generation.

Generator loss function becomes the sum of adversarial loss from (2) and $H$:

$$L_{ACGAN}(G(z)) = L(G(z)) + H(D_c(G(z)) \mid x, c') \quad (5)$$

## 2. Goal of this project

The promise of the GAN is to create realistic images, which sounds like fun to us – why not create a few renderings of the objects that did not exist before? Also, this might allow for easy creation of assets for video games – a huge industry that would be able to pay for such a service. While there are open source implementations, they are rarely engineered to work on anything else but the same dataset as the one authors have worked on. This calls for having own understanding, intuition and a set of primitives before embarking on a more ambitious project with GANs.

This project is centered around reproducing work of others, while creating a software that is maintainable, extendable and configurable.

We setup experiments with having GAN generate images from different small datasets, and see how complexity of the dataset affects the behavior and convergence properties of the model. The results are analyzed according to the 3 criteria defined above – visual quality, diversity and control over the kind of the image produced.

## 3. Methods

### 3.1. Programming language, libraries, hardware

As the basis for the further developments a source code of an example implementation of ACGAN in keras was taken.

Keras is a high-level library in python programming language that provides a set of primitives and objects for Deep Learning with Neural Networks. Two main classes, which contribute the most to the ease of using keras, are Layer class and Model class. Layer is an abstraction for a layer of artificial Neural Network, capable of determining the needed input dimensionality automatically (upon model construction, not changing at the runtime). Specific implementations of Keras layers cover Dense layers, Dropout, Convolutional layers (1D, 2D, 3D), Recurrent layers (LSTM, GRU). Keras works on top of a list of possible "backends" – other libraries that handle low level operations efficiently. As of time of writing, it supports Tensorflow, Theano, CNTK. Tensorflow is the default choice, as it is easy to setup to run on the GPU and offers the best level of integration with Keras (both libraries are developed by google, with tensorflow being initially a google project, and keras being named official high level API later).

In this work we have used Tensorflow as the backend for keras, running on machines with Graphical Processing Unit (GPU). Tensorflow is a framework and API which allows various numerical computations. It is written with production in mind, supporting parallelization of many operations, different hardware it can run on, including distributed solutions. Given a recent NVIDIA GPU and correct CUDA drivers installed, it can execute computation on GPU. In this work, experiments were done on Windows 10 Machines with GTX1080 and GTX1070Ti Graphical Processing Units. It is usual for the training of convolutional networks to gain 10x speedup and more when comparing CPU and GPU training, making it almost a must if multiple models are to be trained and compared.

### 3.2. Datasets

3 Datasets were used to analyze the effects of dataset complexity on the training process and resulting generator model quality. The examples of the images from the datasets used can be seen in the figures 1, 2, 3.

**MNIST**

MNIST [9] is possibly the most used dataset in machine learning. It consists of 60k handwritten digits from 0 to 9, making up 10 classes. The digits are black and white – one color channel in resolution 28 pixel by 28 pixel. It is widely used to demonstrate how Machine Learning works.

There is an opinion voiced by some researchers that MNIST is a very easy dataset, and this makes it a bad test bed for many cases, where models would be able to perform on MNIST, but not in the real life applications. For this reason, Zalando Research came up with Fashion-MNIST, another dataset that is compatible in its format to MNIST.

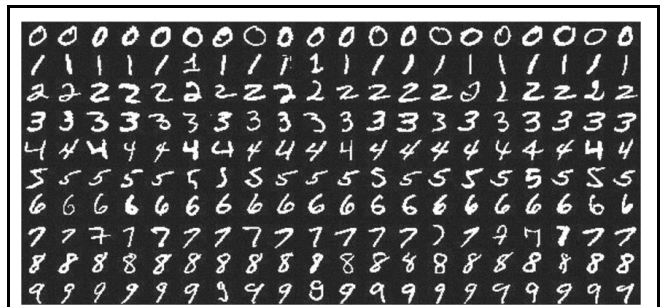

Fig. 1 MNIST dataset examples

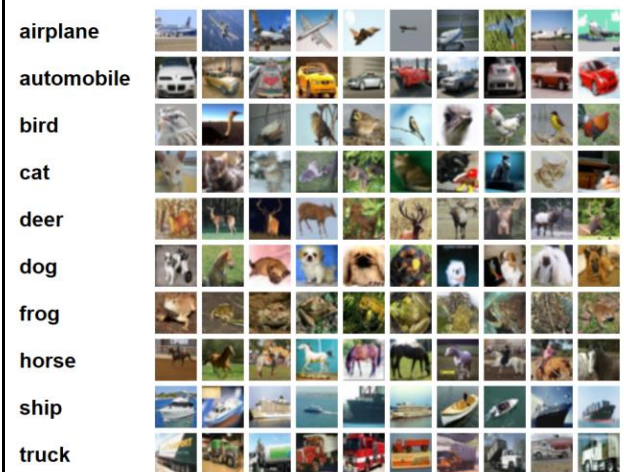

Fig. 2 Fashion-MNIST dataset examples

Fig. 3 CIFAR10 dataset examples

**Fashion-MNIST [8]**

Fashion-MNIST is a more difficult dataset that is made in exactly the same format: 60k images with one color channel and resolution 28x28 pixel. In most applications, it is possible to simply plug it in place of normal MNIST and have your model immediately tested against it. It consists of 10 classes of fashion articles, such as shoes, t-shirts, bags, and others.

CIFAR10

CIFAR10 dataset[10] contains a colored (3 channels, RGB) images with resolution of 32x32 pixels of live and mechanical objects. There are 10 classes in total: 6 animals: bird, cat, deer, dog, frog, horse and 4 inanimate classes: airplane, automobile, ship, truck. These are actual pictures of 3D objects taken from various perspectives. Results of our experiments shows that this dataset is the hardest to generate out of the three.

Networks architecture used

All experiments were run with same architecture of the two networks, generator and discriminator.

Internal structure of the networks used is described next. For the values used following convention is followed: meaning of the vector/tensor *Letter to denote it* [dimensions].

Generator network has internal structure:

Inputs: gaussian noise $z$ [100], class label $c$ [10]

-> Embeddings of the class label $E$ [100]

-> Hadamard Product (z, E) $H$ [100]

-> Dense layer $D1$ [3 * 3 * 384]

-> Reshape $R$ [3, 3, 384]

-> Conv2D $C1$ [7, 7, 256]

-> Conv2D $C2$ [14, 14, 128]

-> Conv2D $C3$ [28, 28, $ch$]

where $ch$ is the number of color channels, 1 for both MNIST and Fashion MNIST, and 3 for CIFAR10. The convolutional layers have used batch normalization between them [4]. The CIFAR10 images are compressed from 32x32 pixels to 28x28 pixels to be able to use the same architecture of the network. All layers use ReLU activation (ReLU(x) = x if x > 0 else 0).

Discriminator has structure as following:

Input: Image $x$ [28, 28, ch]

-> Conv2D C1 [14, 14, 32]

-> Conv2D C2 [14, 14, 64]

-> Conv2D C3 [7, 7, 128]

-> Conv2D C4 [7, 7, 256]

-> Reshape R [7 * 7 * 256 = 12544]

-> Fake_prob D (x) [1] , class label probabilities $D_c(x)$ [10]

The discriminator network uses Leaky ReLU activation, which facilitates relearning on the more dynamic dataset. We observe a degree of similarity in the structures of the networks, only in reverse - the generator starts with a few numbers and up samples them to create an image. The discriminator does the opposite - it shrinks the image to more compact representation with each layer, and finally does classification into the classes and outputs the probability that input image comes from the set of real images *R*.

### 4. Experiments and results

#### 4.1. Metrics

A training of each GAN proceeds in the following manner: overall, 250 epochs (=passes over whole dataset) are made. Batch size of 100 Images were used to train both the generator and the discriminator networks. During the training, following metrics were monitored:

Generation adversarial loss as given by eq. (2),

Generation classification loss as given by eq. (3),

Discriminator adversarial loss as given by eq. (1),

Discriminator classification loss as given by eq. (3).

Also, generated images were saved every 3 epochs to be able to monitor the quality of the produced images and how it developed during the training. As the training is not deterministic, we made 10 separate runs for each dataset.

The plots can be seen in the figures 4, 5, 6, 7. The x axis in the graphs is the number of batches passed. Since the datasets contain 60k images, and batch size was set to 100, it takes 600 batches for the training to make one pass over the dataset (called one epoch). We can observe that for MNIST dataset equilibrium is reached very fast and located around $L_{adv}(D(x))$ = 0.7 for the discriminator and L(G(z)) = 0.775 (the actual values are subject to oscillation). This corresponds to discriminator having accuracy of 50%. Discriminator achieves this by always outputting the probability of image being fake equals to 0.5.

This equilibrium is indeed the predicted state where generator produces images indistinguishable from the original ones by the given discriminator. By analyzing the images (samples from different epochs in Fig.8), we can visually verify that the generated images are of high quality and diverse. They also match the desired class label (0 to 9 per column of images). Therefore, the training has succeeded, and we have achieved all 3 quality criteria mentioned in the section 1.2.

**4.2. Failure modes**

However, for the two other datasets plots of losses look quite different. There are plentiful oscillations in the process, and the trend is for the discriminator to get better, and for generator to get worse on the loss metrics over time. In the Fig. 7 we can see the classification loss of the discriminator. The discriminator network continuously improves at classifying the images from the original datasets, although the datasets obviously present different level of challenge each.

The loss of the discriminator indicates, that for Fashion MNIST dataset it achieved about 82% accuracy at identifying the fake pictures. For the CIFAR10 dataset, this number is around 76%. Visual inspection of the samples for the two datasets (Fig. 9 and 10) show that the generator never achieves visually attractive samples, and also completely lacks diversity within one class. Inspecting the images from the different epochs, we observe the phenomena where the color of the images on average oscillates. Our explanation is that the generator network is unable to grasp the complexity of the objects in the image, and instead uses the training to manipulate statistics of the images (average brightness per color, with just brightness of black and white images for the Fashion MNIST). It also keeps changing the look of the images, preventing the discriminator from ever being 100% certain about how up-to-date fake images look like (as the learning speed is capped by the learning rate).

**4.3. Analyzing the generated images in detail, limited resource reuse hypothesis**

In figures 11, 12, 13, 15 zoom-ins on the generated images are available. Figures 11, 12, 13 come from the epochs where generator already displays oscillatory behavior. In figures 11 and 12 some objects can be recognized (some imagination required). At the same time other objects are barely recognizable. It is important to note that the "well-generated" objects change with the flow of training.

This hints us at potential insight (for now just a hypothesis) to the mechanics of the observed oscillations. As generator is a network with very limited number of filters, it might be actually impossible to embed the complexity of all the objects in those filters at once. Therefore, the network makes the trade-off and maxes out on the realism of a single or few of categories. However, due to statistical nature of the training, soon discriminator will be biased and will start giving low scores for this category independent of the quality and will also specialize in recognizing the fakes in this category. This triggers the shift to another object that can be made realistic now that discriminator has used up resources to be more proficient on the current favored class. As the dataset has color imbalances (most frogs are greenish, planes are on white-blue skies) it can be one of the explanations for the observed oscillations of the color of all the generated images.

From Fig. 15 we can see that in fact the generator attempts to introduce variety into the images when it can be done with a rather small change, so in principle it is justified to expect a stronger generator with more resources to produce better quality of images also on these more difficult datasets.

## 5. Conclusion

Through experiments, we have seen that despite exactly same format, datasets differ in their difficulty merely based on the content of the images. MNIST and Fashion-MNIST use same data format, yet ACGAN network is able to easily generate realistic images from the former but not the latter dataset.

From the literature it is known that bigger networks successfully generate images for much more complex datasets than the Fashion MNIST and CIFAR10, such as e.g. ImageNet with 1000 classes of objects and colored images in the resolution of over 256 by 256 pixels. [3], [5].

It can be concluded that for successful training on these datasets it is necessary to take network with greater capacity, in terms of number of neurons and/or more layers. Also, more advanced architectures like parallel inception layers[6] and residual layers are used in the literature[7].

Further experiments are needed to gain more specific understanding into the nature of complexity of the datasets. By synthesizing images it can be possible to analyse the effect of shape variance, amount of objects per picture and other similar attributes.

By varying single parameters of the generator network, it should be possible to understand which parameter represents the key bottleneck. For example, it is thinkable that a certain depth of the network is necessary, as otherwise the logical hierarchy of the features can not be learned.

## AUTHOR INFORMATION

**Ilya Kamenshchikov,** Machine Learning Engineer at Capgemini Deutschland GmbH and MSc. Bionics student, Faculty of Technology and Bionics, Hochschule Rhein-Waal.

Co-author and supervisor: **Prof. Dr. Matthias Krauledat,** Professor of Informatics at Faculty of Technology and Bionics, Hochschule Rhein-Waal.

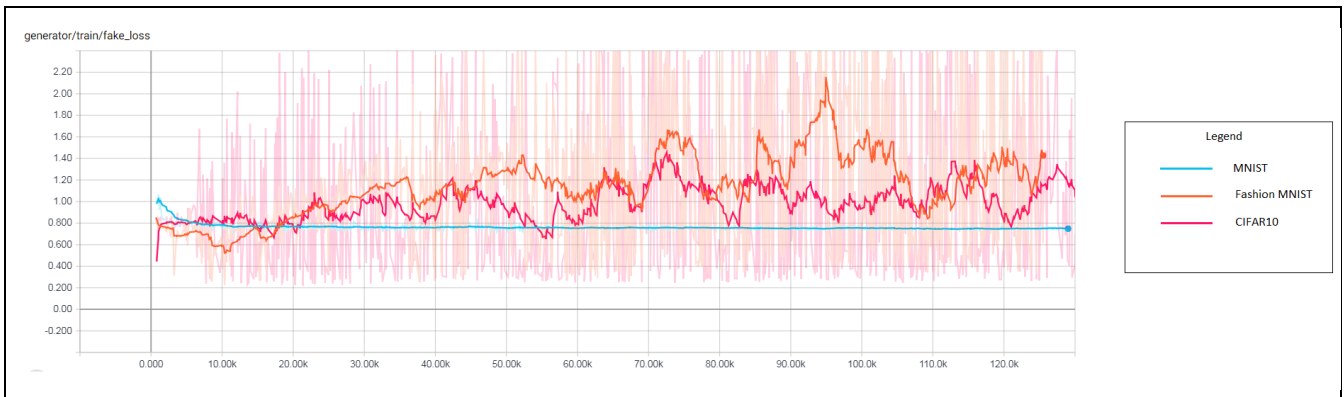

Fig. 4 Generator Adversarial Loss

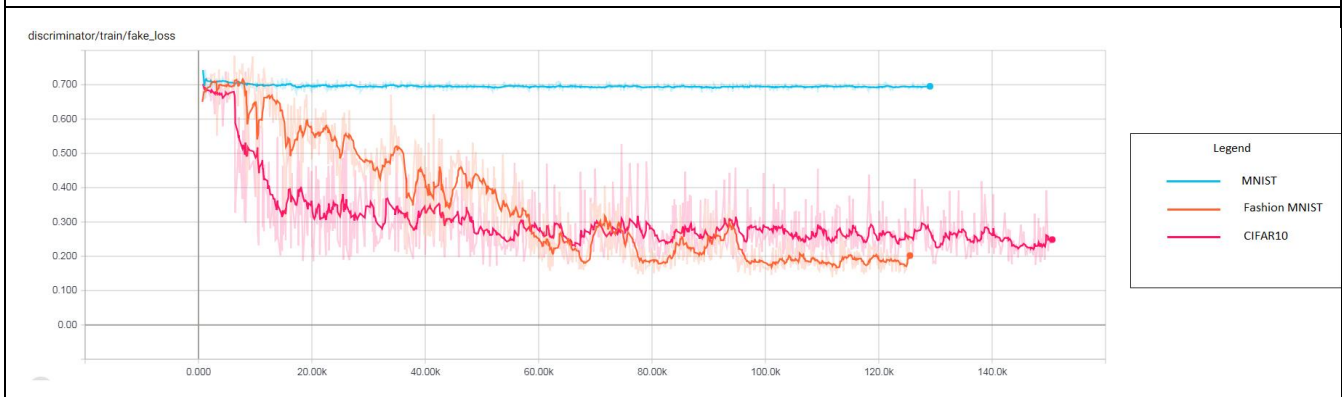

Fig. 5 Discriminator adversarial loss

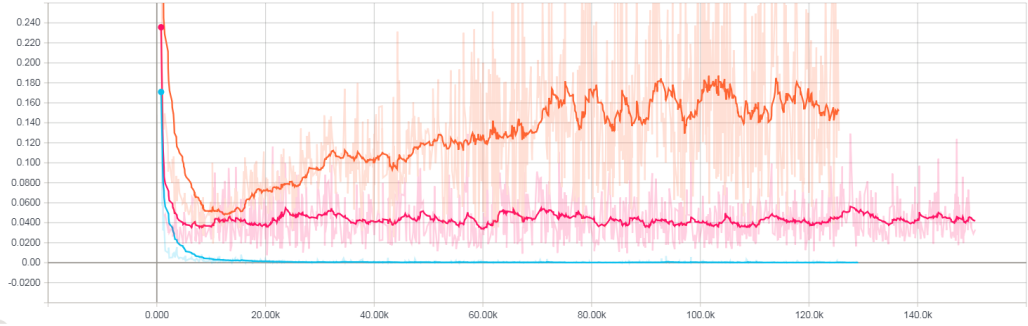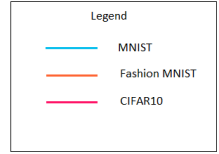

Fig. 6 Generator Classification loss

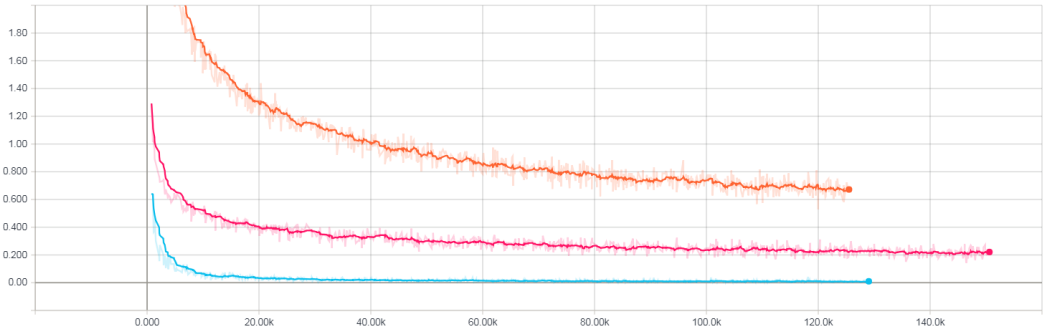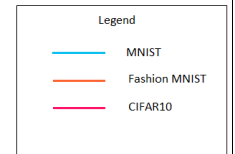

Fig. 7 Discriminator classification loss

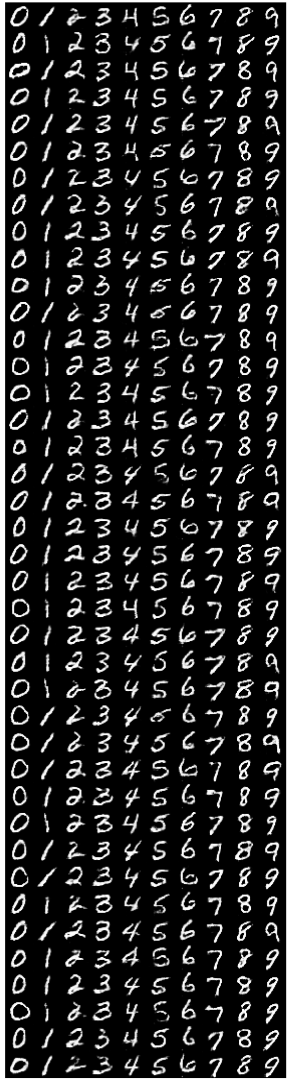 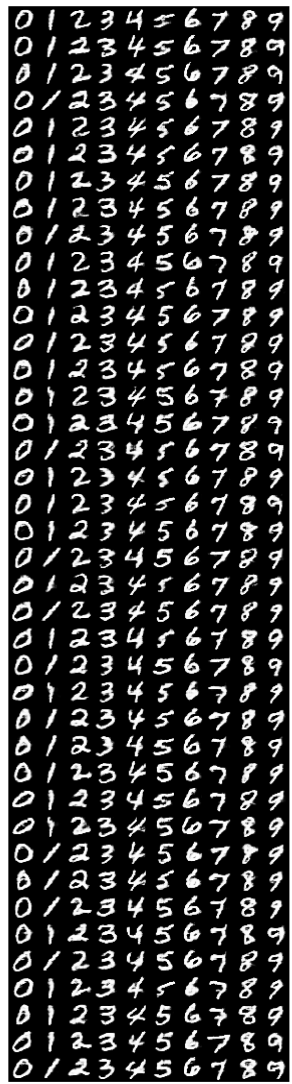 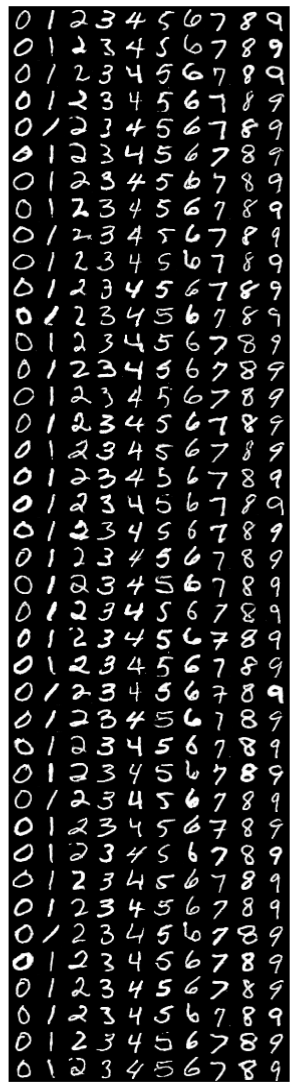 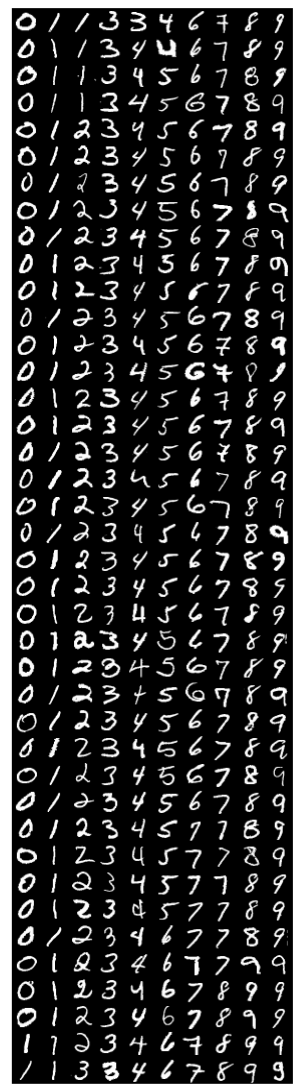

Fig. 8 Generated images for the MNIST dataset and samples of the original dataset

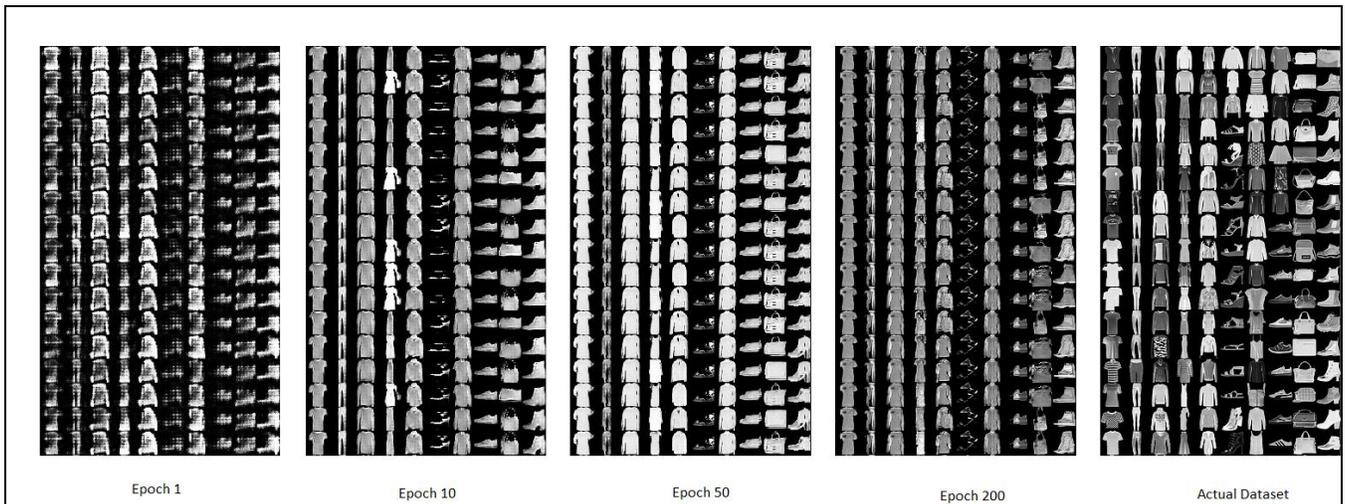

Fig. 9 Generated images for the Fashion MNIST Dataset and samples of the original dataset

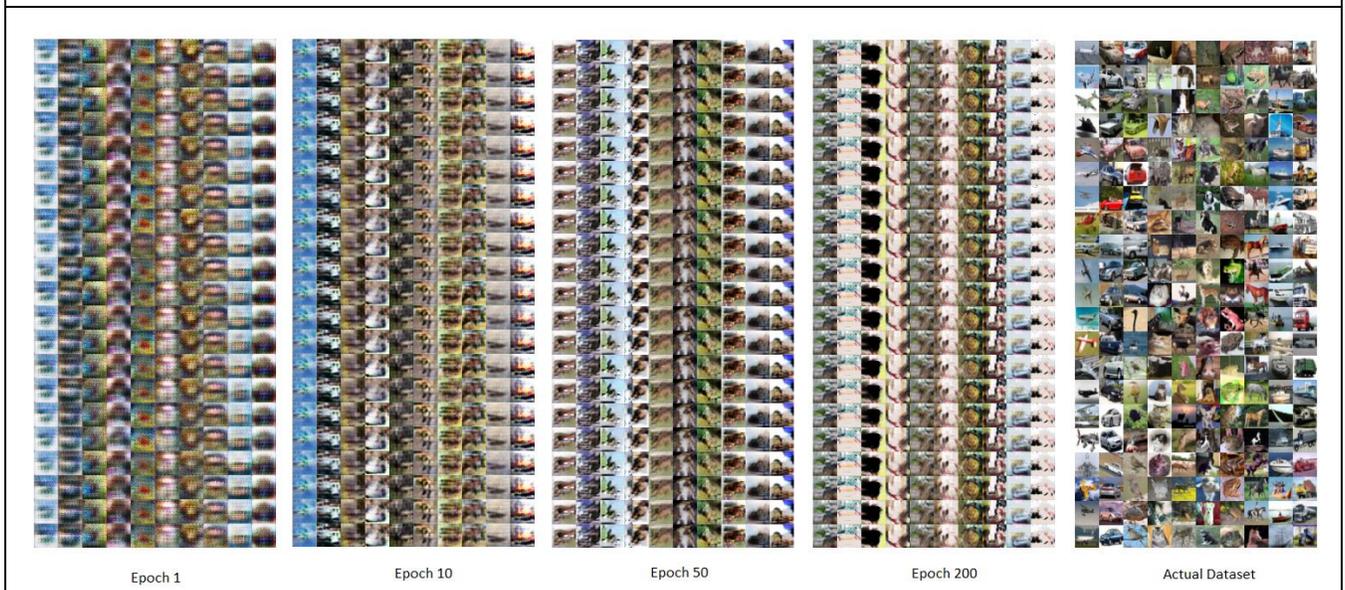

Fig. 10 Generated images for the CIFAR10 dataset and samples of the original dataset

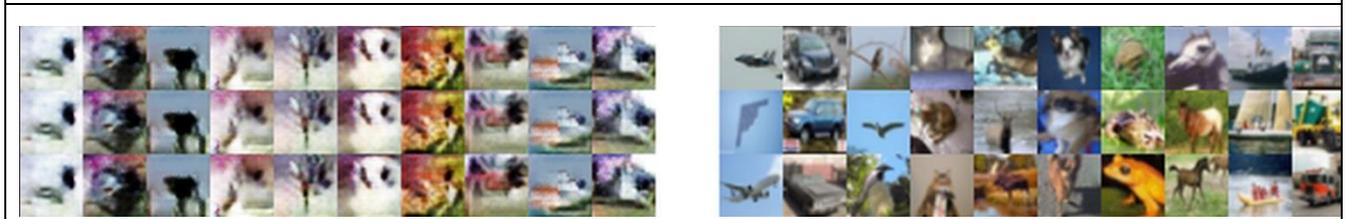

Fig. 11: Cifar – maybe penguin in column 3? Maybe a Pekinese Dog in column 6?

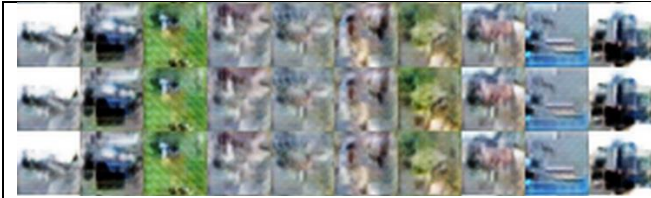 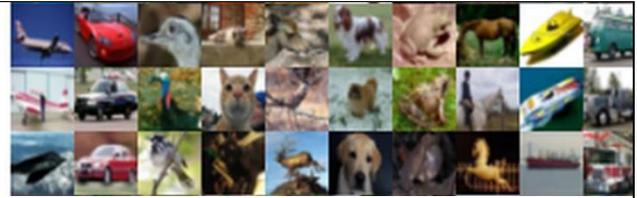

Fig. 12 Cifar – Maybe Horse in column 7

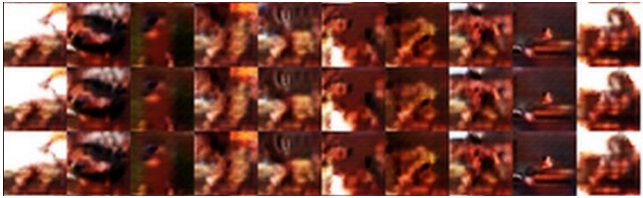 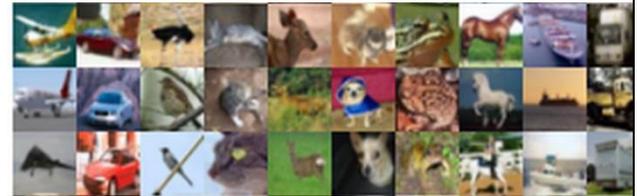

Fig. 13 Cifar – hard to recognize

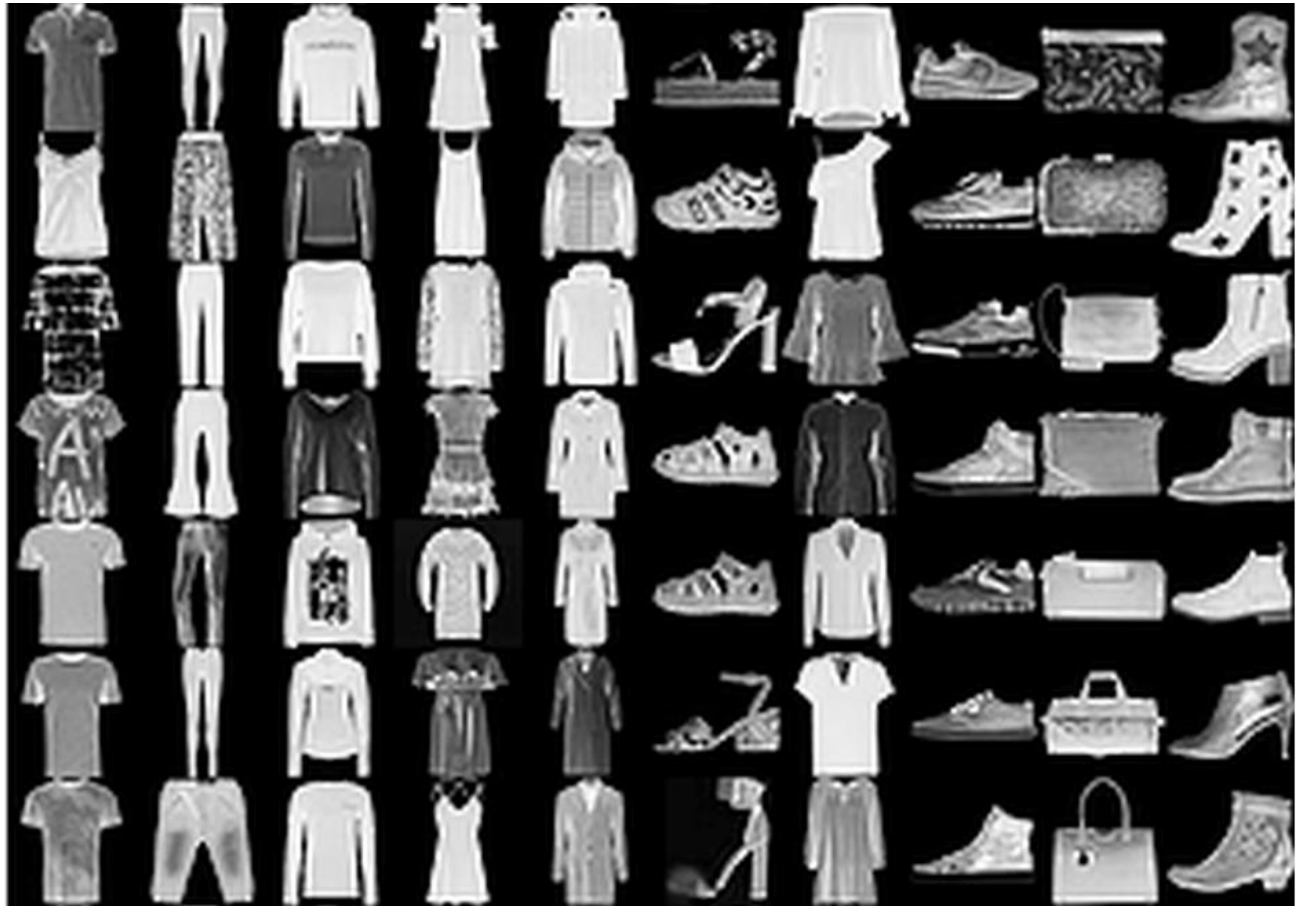

Fig. 14 Fashion-MNIST (Real) zoom-in

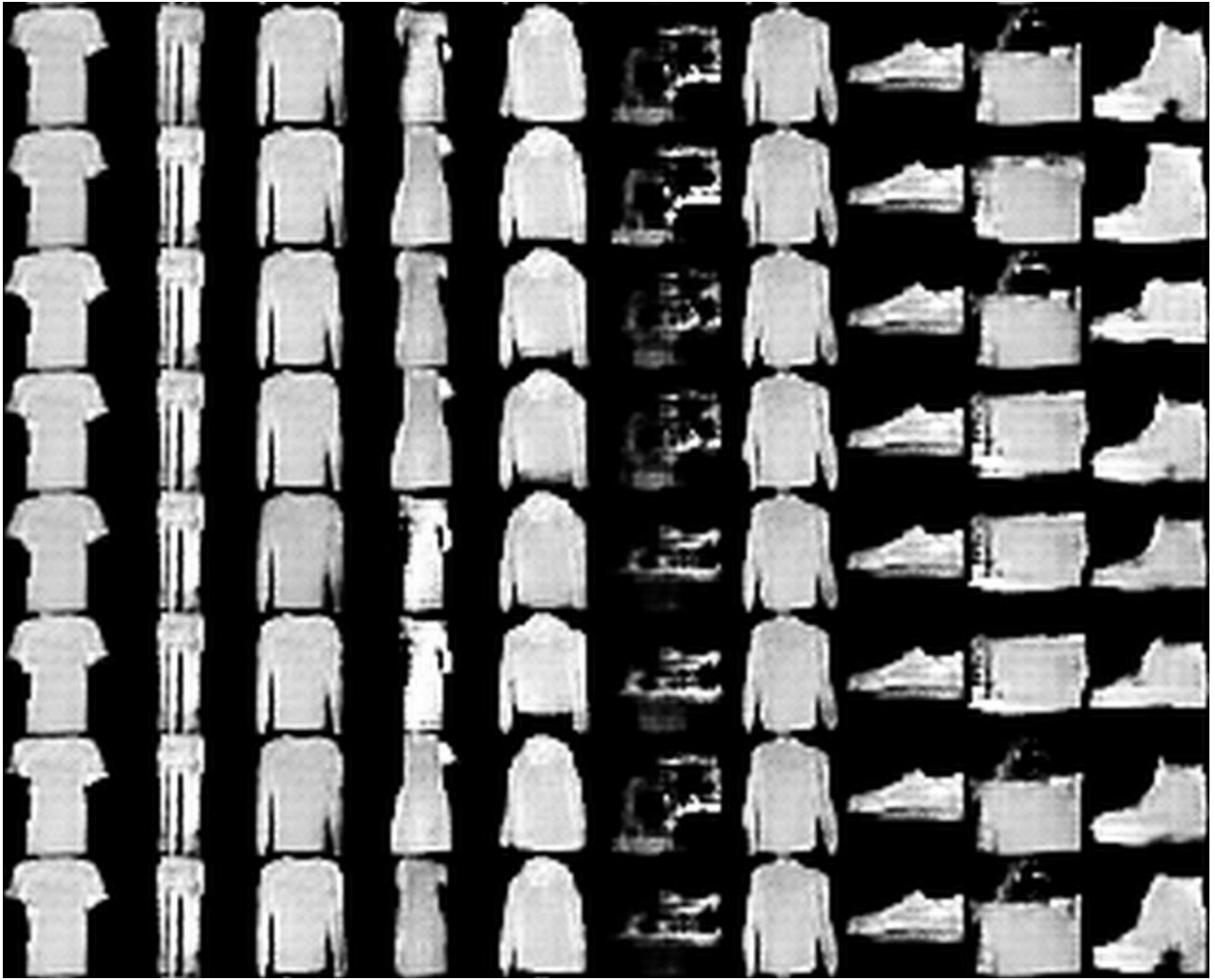

Fig.15 Fashion-MNIST generated zoom-in